\documentclass[sigconf]{acmart}

\usepackage{subfig}
\usepackage{graphicx}
\usepackage{amsmath}
\usepackage{amsfonts}
\usepackage{CJKutf8}

\AtBeginDocument{%
  \providecommand\BibTeX{{%
    \normalfont B\kern-0.5em{\scshape i\kern-0.25em b}\kern-0.8em\TeX}}}

\setcopyright{acmcopyright}
\copyrightyear{2018}
\acmYear{2018}
\acmDOI{10.1145/1122445.1122456}

\copyrightyear{2021}
\acmYear{2021}
\setcopyright{acmcopyright}
\acmConference[SUMAC '21] {Proceedings of the 3rd Workshop on Structuring and Understanding of Multimedia heritAge Contents}{October 20, 2021}{Virtual Event, China.}
\acmBooktitle{Proceedings of the 3rd Workshop on Structuring and Understanding of Multimedia heritAge Contents (SUMAC '21), October 20, 2021, Virtual Event, China}
\acmPrice{15.00}
\acmISBN{978-1-4503-8668-5/21/10}
\acmDOI{10.1145/3475720.3484441.}
\begin{document}
\fancyhead{}

\title{Built Year Prediction from Buddha Face with Heterogeneous Labels}

\author{Yiming Qian}
\email{yimingqian@ids.osaka-u.ac.jp}
\orcid{0000-0002-1795-2038}
\affiliation{%
  \institution{Osaka University}
  \city{Osaka}
  \country{Japan}
}

\author{Cheikh Brahim El Vaigh}
\email{cheikh-brahim.el-vaigh@irisa.fr}
\affiliation{%
  \institution{Univ. Rennes, CNRS, IRISA}
  \city{Lannion}
  \country{France}}

\author{Yuta Nakashima}
\email{n-yuta@ids.osaka-u.ac.jp}
\affiliation{%
  \institution{Osaka University}
  \city{Osaka}
  \country{Japan}
}

\author{Benjamin Renoust}
\email{renoust@ids.osaka-u.ac.jp}
\affiliation{%
 \institution{Median Technologies, and Osaka
University}
 \city{Valbonne}
 \country{France}
 }

\author{Hajime Nagahara}
\email{nagahara@ids.osaka-u.ac.jp}
\affiliation{%
  \institution{Osaka
University}
  \city{Osaka}
  \country{Japan}}

\author{Yutaka Fujioka}
\email{fujioka@let.osaka-u.ac.jp}
\affiliation{%
  \institution{Osaka
University}
  \city{Osaka}
  \country{Japan}}

\renewcommand{\shortauthors}{Anonymous Submitter}
\newcommand{\CBEL}[1]{\textcolor{cyan}{CBEL:~#1}}
\newcommand{\YQ}[1]{\textcolor{red}{YQ:~#1}}

\begin{abstract}
Buddha statues are a part of human culture, especially of the Asia area, and they have been alongside human civilisation for more than 2,000 years. As history goes by, due to wars, natural disasters, and other reasons, the records that show the built years of Buddha statues went missing, which makes it an immense work for historians to estimate the built years. In this paper, we pursue the idea of building a neural network model that automatically estimates the built years of Buddha statues based only on their face images. Our model uses a loss function that consists of three terms: an MSE loss that provides the basis for built year estimation; a KL divergence-based loss that handles the samples with both an exact built year and a possible range of built years (e.g., dynasty or centuries) estimated by historians; finally a regularisation that utilises both labelled and unlabelled samples based on manifold assumption. By combining those three terms in the training process, we show that our method is able to estimate built years for given images with 37.5 years of a mean absolute error on the test set.
 
\end{abstract}




\maketitle

\section{Introduction}
Buddhism has started in India and spread all over the Asian subcontinent from India to, e.g.~Japan through China. As Buddhism flourished along the centuries within several civilisations and dynasties, people started embracing the new religion, and now it is considered as the fourth largest religion in the world. They created their own Buddha statues, which are not only religion symbols but also art pieces that express their own culture. 

The built information will help historians to connect the design of Buddha with the historical events such as dynasty change, revolution, etc. However, as history goes by, due to wars, natural disasters, and other reasons, the document that records the history of Buddha statues went missing. 

It is an immense work for historians to estimate the built year of statues after the record is lost. Chemical analysis-based methods, such as radiocarbon dating \cite{taylor2016radiocarbon}, weathering-base dating~\cite{purdy1987weathering}, thermoluminescence dating \cite{wintle1982thermoluminescence}, can be used to estimate Buddha statues' built years, but these methods have some drawbacks. For example, radiocarbon dating is only capable of identifying the date of organic components, which is not always the case for Buddha statues. The weathering-based method analyses the degrading of the surface due to weathering, being applicable to statues in limited environmental settings. Thermoluminescence dating is only applicable to statues made of a material that was heated during the building process, such as bronze, ceramic, or gold coating. Those methods are expensive, time consuming, and only applicable to limited situations. 

In this work, we propose a method that estimates a Buddha statues built year leveraging only the image of Buddha faces. \textit{To the best of our knowledge, we are the first to address this task in an automatic manner}. We use the dataset of Buddha statues presented by Renoust et al.~\cite{renoust2019historical}, which, to the best of our knowledge, is the only dataset that comes with a rich set of annotations on built time, materials, etc. One major challenge in this dataset is its heterogeneous labels; that is, due to the inherent nature of cultural heritage as mentioned above, Buddha statue history can be lost, and thus some labels are completely missing or can only be roughly estimated by historians. As for their built time, many samples do not have relevant labels at all; some come with their exact built year, and the others only have more ambiguous expressions on the built time, such as the dynasty or century that they were established in.


In this paper, we are proposing a method that estimates the built year from Buddha statue's facial images using a deep learning-based model. We trained a regression model to compute the built year from the image embedding extracted with a convolutional neural network (CNN) backbone. To make full use of the heterogeneous labels, we take advantage of weakly-labelled and unlabelled images in a semi-supervised manner and devise a loss function so that labels with different ambiguities can be incorporated into training. More specifically, our loss function consists of a MSE loss to reduce an image embedding to a scalar representing the built year, a Kullback-Leibler  (KL) divergence-based loss that constraints the built year values to a certain given range, and a regularisation loss designed to unleash the information hidden in the unlabelled images.

\textbf{Contribution}. In addition to the above mentioned new loss functions, we propose to represent the built time as Gaussian functions. This unifies the labels in different ambiguity levels and provides a way to incorporate them in our KL divergence-based loss term. We experimentally show that our method can estimate the built year with 37.5 years of mean absolute error (MAE). 



\section{Prior Work}
\label{relatedwork}

In this section, we limit the discussion to the semi-supervised learning framework (Section \ref{relatedwork:ssl}) and how it can be used to address the fine-grained task of Buddha statue's built year prediction (Section~\ref{relatedwork:artanalysis}).

\subsection{Built Year Prediction}
\label{relatedwork:artanalysis}
When literature about the Buddha statue is missing, their built year can be identified using some chemical analysis-based method, which analyses the chemical components of the material. Such methods are both time consuming and material-dependent since they can only be applied for some specific materials, such as wood or stone, in most cases demanding special equipment. The recent movement of digitisation of the humanities allows to build digital scans of Buddha statues, opening the door to automatic analysis that requires data and computers.

Machine learning, or more specifically CNNs, such as ResNet \cite{resnet50} and VGG \cite{vgg16}, has been one of the main driving forces of this movement, allowing to automatically extract features from an image that give accurate representation for different kinds of entities, such as text, natural images, or art pieces. They are extensively used as off-the-shelf model in styles classification~\cite{GCNBoost,garcia2019context,renoust2019historical,bar2014classification}, authorship identification~\cite{ma2017part}, and artworks retrieval~\cite{,mao2017deepart}. 
Mensink~\cite{mensink2014rijksmuseum} proposed a method that applies max-margin regressor to SIFT features to identify painting built years. Strezoski~\cite{strezoski2018omniart} improved this built year estimation task with a multitask learning deep learning network. Classification of various aspects of Buddha statues is addressed in \cite{renoust2019historical,GCNBoost}, but their built years are predicted in the century basis, which only gives rough ideas about their establishment. The major drawback of the traditional deep learning approach is it requires an enormous amount of labelled data during the training process. Furthermore, the past approaches require built year labels in normalized form, which often leads to quantizate the label into centuries and reformulate the prediction problem as a classification problem.

In our Buddha project, we do not have the luxury of having a large labelled dataset. To handle this situation, we start to investigate the possibility of applying semi-supervised learning to relax this labelled data shortage. 

\subsection{Semi-supervised Learning}
\label{relatedwork:ssl}
Semi-supervised learning is a way to combine labelled data and large amounts of unlabelled data into the same training process. In this context, several assumptions are adopted to make the best of unlabelled data. 

The most widely used one is the cluster assumption, stating that: \textit{if two samples are in the same cluster, they are likely to belong to the same class}. This is a strong assumption particularly for classification problems, which allows to give pseudo labels to unlabelled samples. In this way, the labelled data and pseudo-labelled data can be used in the training process. With inclusion of pseudo-labelled data, the training data pool significantly increases, which improves the data diversity and lowers the chance of over-fitting. There are two major ways of semi-supervised learning: transductive learning \cite{chen2020simple,moscovich2017minimax} and inductive learning \cite{sohn2020fixmatch,tarvainen2017mean,zou2019confidence,berthelot2019mixmatch,li2019learning}. Transductive learning produces labels only for unlabelled data available in the training, while the inductive learning assigns new data with a label from prediction. 

The manifold assumption is another popular assumption, which is applicable for both regression and classification problems \cite{10.5555/2981562.2981663,li2021multi,berikov2021solving,rwebangira2009local,zhu2003semi,jeanxieermon_ssdkl_2018}. It states that: \textit{the high-dimensional data lie roughly on a low-dimensional manifold}, which implies that in the manifold the densely sampled regions are smoother with smaller gradient. Under this assumption, a regularizer with radial basis function kernel can be used to enforce the smoothness among label and unlabelled data~\cite{berikov2021solving}.

Our paper devises a semi-supervised regression model that predicts built years with high accuracy. Furthermore, we study how unlabelled data facilitate the regression task, taking into account the manifold assumptions in the context of semi-supervised learning.

\section{Dataset} 
We obtained the dataset from Renoust et al.~\cite{renoust2019historical}, which consists of 7,518 scanned Buddha images from 5 different books. The retina face detection algorithm \cite{deng2019retinaface} was deployed by the authors to extract the face of statues. These face images were then aligned and resize to $112\times112$, following the same process in \cite{deng2019arcface}. The algorithm found 4,949 Buddha face images in the dataset. Among them, only 1,887 have built time labels, while the remaining do not provide any information about their built time. The built time labels associated with respective images fall into three types: \textit{dynasty} in which the statue was built, ranging from 40 years to 700 years, \textit{century}, and exact \textit{year}. Hereinafter, we call these labels \textit{built time} collectively, while using \textit{built year} whenever it pinpoints a certain year. We picked out the label with the smallest range if multiple labels are available. The distribution of the labels is as follows:
\begin{itemize}
	\item \textit{dynasty}: 320 samples
	\item \textit{century}: 316 samples
	\item \textit{year}: 1,251 samples
\end{itemize}

\begin{figure}[t]
	\centering
	\subfloat{\includegraphics[width=7cm]{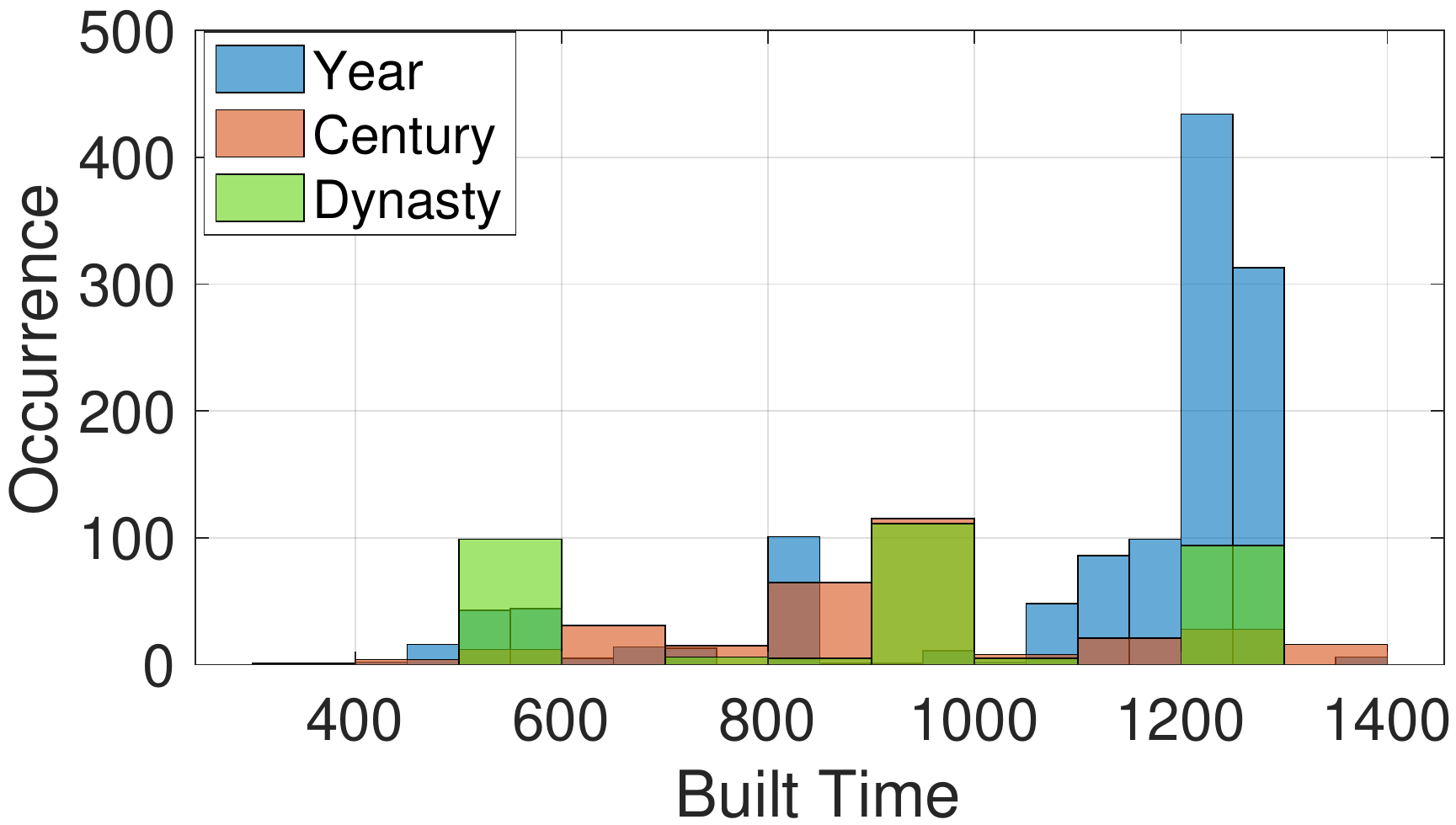}}
	\caption{Distribution of built time on labelled data.}
	\label{fig:builtYearDistribution}
\end{figure}

We randomly split the dataset into 70\% (i.e., 3,464 samples) for training and 30\% (i.e., 1,485 samples) for testing, where 1,340 out of 3,464 samples have built time labels in the training set and 547 out of 1,485 samples in the testing set.

\begin{figure}[htpb!]
	\centering
	\subfloat[]{\includegraphics[width=2.6cm]{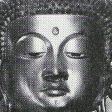}}\thinspace
	\subfloat[]{\includegraphics[width=2.6cm]{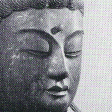}}\thinspace
	\subfloat[]{\includegraphics[width=2.6cm]{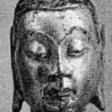}}\\
	\subfloat[]{\includegraphics[width=2.6cm]{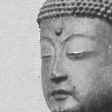}}\thinspace
	\subfloat[]{\includegraphics[width=2.6cm]{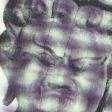}}\thinspace
	\subfloat[]{\includegraphics[width=2.6cm]{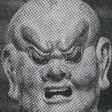}}\\
	\subfloat[]{\includegraphics[width=2.6cm]{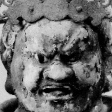}}\thinspace
	\subfloat[]{\includegraphics[width=2.6cm]{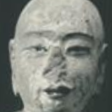}}\thinspace
	\subfloat[]{\includegraphics[width=2.6cm]{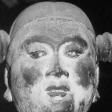}}\thinspace
	\begin{CJK}{UTF8}{min}
	\caption{Some samples of Buddha face images in the dataset. (a) Amidanyorai (阿弥陀如来), (b) Seishibosatsu (勢至菩薩), (c) Mirokubutsu (弥勒仏), (d) Amidanyorai (阿弥陀如来), (e) Fud\=omy\=o\=o (不動明王) (f) Kong\=orikishi (金剛力士) (g) Z\=och\=oten (増長天) (h) Jisha (侍者) (i) Zenzaid\=oji (善財童子).}
	\label{fig:buddhafacesamples}
	\end{CJK}
\end{figure}

Figure \ref{fig:buddhafacesamples} shows some sample Buddha face images after alignment. Many of them have missing facial parts (Fig.~\ref{fig:buddhafacesamples} (h)). All images of the dataset were collected by scanning printed books, which often leads to large colour distortion from the original statues (Fig.~\ref{fig:buddhafacesamples} (c)). The original images in the books were captured by multiple cameras and their intrinsic parameters are unknown. The amount of object distortion caused by different cameras' focal lengths and capturing distances across images make the task particularly challenging. For example, there are different types of Buddha statues as shown in \ref{fig:buddhafacesamples}; there can be multiple Buddha statues of Amidanyorai, even from the same authors but established at different periods of time, and they only have a slight differences in their faces. Moreover, the original books come with artefacts due to the AM and FM screening halftone printing process (Fig.~\ref{fig:buddhafacesamples} (f)). The scanners used in the digitisation process can also introduce noises and artefacts known as Moir\'{e} patterns (Fig.~\ref{fig:buddhafacesamples} (e)).

\section{Method} 

Our task is to estimate the built year $t$ of a given image $x$. Although the transductive paradigm may also work for our task because the number of cultural heritages merely increases, we this time choose the inductive paradigm. More specifically, we denote our \emph{labelled} training set $\mathcal{D}_\text{L} = \{(x_i, t_i) \mid i = 0, \dots, I_\text{L}\}$ and \emph{unlabelled} training set $\mathcal{D}_\text{U} = \{x_i \mid i = 0, \dots, I_\text{U}\}$ for training, where $x_i$ and $t_i$ are the $i$-th image and the corresponding label; and $I_\text{L}$ and $I_\text{U}$ is the numbers of samples in the labelled  and unlabelled training sets, respectively. 

\begin{figure}[t]
	\centering
	\subfloat{\includegraphics[width=7cm]{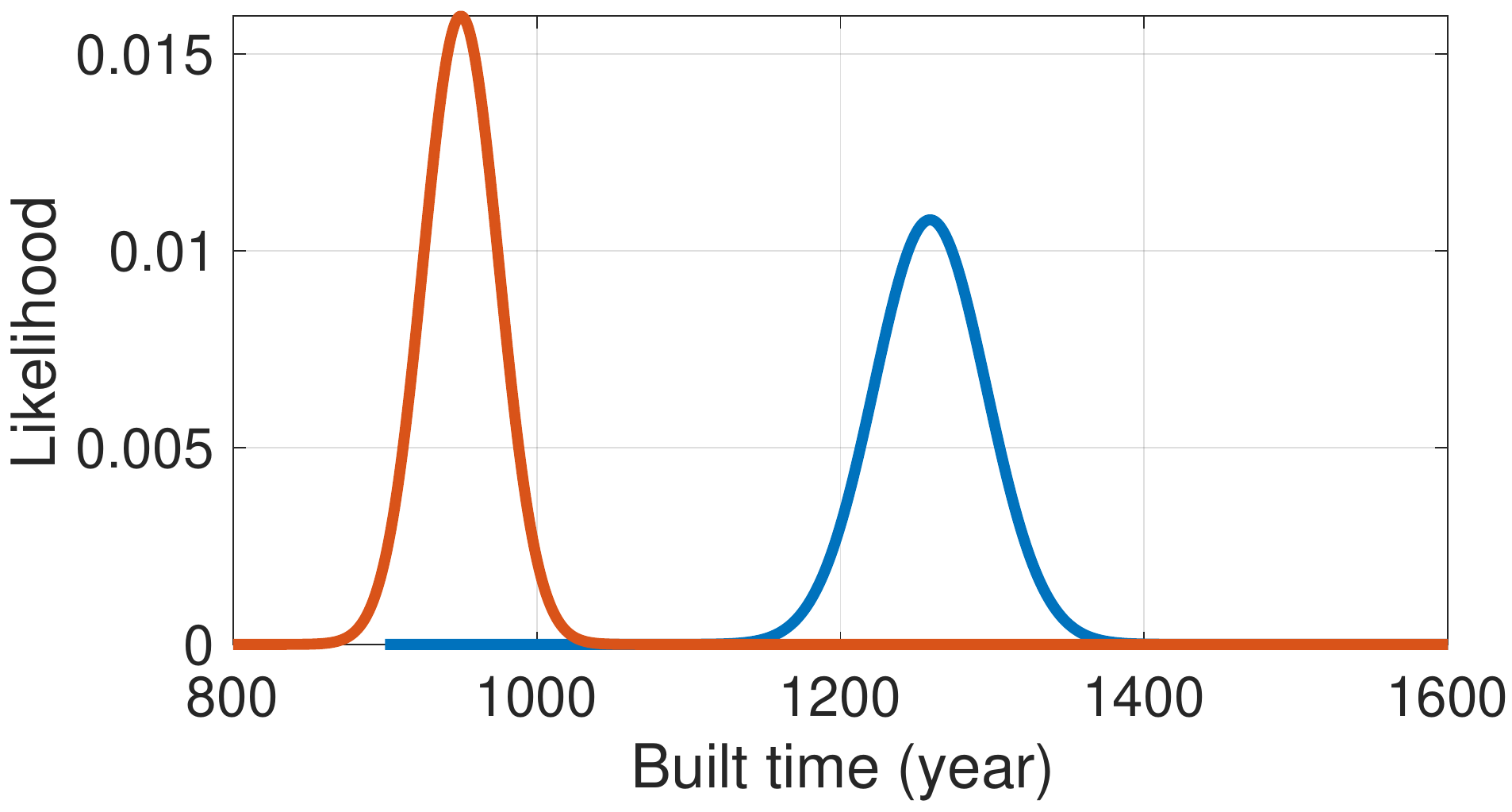}}
	\caption{Two examples Gaussian to represent built times. The blue line is $\mathcal{N}(1259, 37^2)$, which represents the Kamakura dynasty (1185--1333). The red line is $\mathcal{N}(950, 25^2)$, covering the 10th century.}
	\label{fig:builtyearlabels}
\end{figure}

As mentioned previously, $\mathcal{D}_\text{L}$ contains three types of labels, i.e., one that gives exact \textit{year} of building and the others that give the range of built time (\textit{dynasty} or \textit{century}). We denote the corresponding sets by $\mathcal{D}_\text{L}^\text{Y}$, $\mathcal{D}_\text{L}^\text{D}$, and $\mathcal{D}_\text{L}^\text{C}$, respectively ($\mathcal{D}_\text{L} = \mathcal{D}_\text{L}^\text{Y} \cup \mathcal{D}_\text{L}^\text{D} \cup \mathcal{D}_\text{L}^\text{C}$).  In order to unify labels with different ambiguities, we represent the built time with Gaussian $\mathcal{N}(\mu, \sigma^2)$. For samples with exact built year, we use the built year as mean $\mu$ and 2.5 as standard deviation $\sigma$, covering 10 years centred at the year in the 95\% confidence interval. A century spans 100 years, so the mean is set to the middle of the century (i.e., $\mu = 1,450$ for the 15th century) and 25 years as standard deviation. For dynasties,  we use the middle year of the dynasty period as $\mu$ and the quarter of the dynasty period as $\sigma$, which covers the dynasty within the 95\% confidence interval. Therefore, our label $t$ is given by a tuple $t = (\mu_t, \sigma_t)$, which identifies a Gaussian. Figure \ref{fig:builtyearlabels} illustrates two examples Gaussian. The blue line represents the \textit{Kamakura} dynasty, which starts at 1185 and lasts until 1333. The mean for this dynasty is 1259 and the standard deviation is 37. The red line is the 10th century (from 901 to 1000), where the mean and standard deviation are 950 and 25, respectively.

Our model consists of a ResNet50 variant of Arcface \cite{deng2019arcface}, pre-trained with 17 million human face datasets \cite{an2020partical_fc}, as backbone, of which the last batch normalisation layer is connected to a fully connected layer to predict built year. 

To make full use of the dataset with labelled and unlabelled images, we design a dedicated semi-supervised loss function, consisting of three terms, which are the mean squared error (MSE) loss, a KL divergence-based loss to better incorporate the different ambiguity levels, and a regularisation to make a smoother manifold with unlabelled samples. 

\textbf{MSE loss.} 
The MSE loss provides basic supervision merely for samples with exact built years. Let $f(x)$ denote our model to regress the built year from image $x$. Our MSE loss $E$ is defined as:
\begin{equation}
    E=\frac{1}{|\mathcal{D}_\text{L}^\text{Y}|}\sum_{(x, t) \in \mathcal{D}_\text{L}^\text{Y}} \| f(x)-\mu_t  \|_2,
\end{equation}
where $\mu_t$ is the mean of $t$. 

\textbf{KL divergence-based loss.} This loss utilises the different ranges of the built time labels in the dataset. Intuitively, an arbitrary pair of predicted built year must have a similar relationship (i.e., the closeness) to the ground-truth built time, which is not as straightforward as computing the distance between two samples, since the built time labels come with different ambiguities. Inspired by t-SNE \cite{van2008visualizing}, we encode such pairwise relationships into conditional probability $q(f(x')\mid f(x))$ of prediction $f(x')$ given $f(x)$ (or conditional probability $p(t' \mid t)$ of ground-truth $t'$ given $t$), where $(x, t)$ and $(x', t')$ are in $\mathcal{D}_\text{L}$. We enforce $q$ being similar to $p$, so that the pairwise relationships in the ground-truth built year labels are maintained in the predictions.

More specifically, for prediction, we define the conditional probability $q(f(x') \mid f(x))$ as the likelihood of $f(x')$ assuming that it follows $N(f(x), \sigma_t^2)$, normalised over all samples in all training samples, where $\sigma_t$ is borrowed from the ground truth-label $t$ associated with $x$ (i.e., $(x, t) \in \mathcal{D}_\text{L}$). This can be formalised as:
\begin{equation}
\label{eqn:qvalue}
    q(f(x') \mid f(x)) = \frac{\frac{1}{\sigma_t\sqrt{2\pi}} \exp\left\{-\frac{(f(x')-f(x))^2}{2\sigma_t^2}\right\}}
    {\sum_{x'}\frac{1}{\sigma_t\sqrt{2\pi}} \exp\left\{-\frac{(f(x')-f(x))^2}{2\sigma_t^2}\right\}}.
\end{equation}
Similarly, we can define the pairwise conditional probability $p(t' \mid t)$ from ground-truth labels $t$.
\begin{equation}
\label{eqn:pvalue}
    p(t' \mid t)= \frac{\frac{1}{\sigma_t\sqrt{2\pi}} \exp\left\{-\frac{1}{2}\frac{(\mu_{t'}-\mu_t)^2}{\sigma_t^2}\right\}}{\sum_{t'}\frac{1}{\sigma_t\sqrt{2\pi}} \exp\left\{-\frac{1}{2}\frac{(\mu_{t'}-\mu_t)^2}{\sigma_t^2}\right\}}.
\end{equation}
The summations in the above two equations are computed over $(x', t') \in \mathcal{D}_\text{L}$.

Our KL divergence-based loss $C$ based on conditional probability $q$ and $p$ is calculated by
\begin{equation}
	C =\sum_{(x, t)} \text{KL}(p \| q)=\sum_{(x, t)} \sum_{(x', t')} p(t' \mid t) \log\left( \frac{p(t' \mid t)}{q(f(x') \mid f(x))}\right),
\end{equation}
where $(x, t)$ and $(x', t')$ are in $\mathcal{D}_\text{L}$.

\textbf{Regularisation.} Our regularisation loss is designed to unleash the information that is hidden in the unlabelled samples. It uses the manifold assumption, enforcing a smoother manifold by incorporating unlabelled data in the training process. 

According to the manifold assumption, feature vectors from the backbone lie on a smooth low-dimensional manifold in the feature space. Berikov and Litvinenko \cite{berikov2019semi} suggested that the decision function should change slowly in regions where feature vectors are densely distributed. From this assumption, the large amount of unlabelled data can help smooth the manifold and so regularise the training process. We adapted the radial basis function (RBF)  \cite{berikov2019semi,rwebangira2009local,zhu2003semi} to approximate the smoothness of the manifold. 

Let $g(x)$ denote feature vector $v$ from our backbone $g$, where our model $f$ is the composition of $g$ and regressor (a fully connected layer) $h$ (i.e., $f(x) = h(g(x))$. Following \cite{berikov2019semi}, the pairwise regularisation weight $\phi(g(x),g(x'))$ is applied between a pair of a labelled sample and an unlabelled sample, which is given by:
\begin{equation}
    \phi(g(x),g(x'))=\exp\left\{-\frac{ \| g(x)-g(x') \|^2}{2l^2}\right\},
\end{equation}
where $g(x)$ and $g(x')$ are the output from the last batch-normalisation layer, and $l$ is a parameter to control the smoothness. We used $l = 0.75$ based on \cite{berikov2019semi}. The regularisation term $R$, which is the mean of the regularisation weights, is given as follows:
\begin{equation}
    R = \frac{1}{|\mathcal{D}_\text{L}| \; |\mathcal{D}_\text{U}|} \sum_{x} \sum_{x'} \phi(g(x),g(x')),
\end{equation}
where the first and second summations are computed over the images $x$ and $x'$ in $\mathcal{D}_\text{L}$ and $\mathcal{D}_\text{U}$, respectively. This term enlarges data pools used in training by 200\%, allowing our model to be robust to overfitting. 

\textbf{Overall loss function.} Our overall loss function $\ell$ is a linear combination of three terms, given by 
\begin{equation}
    \ell=\alpha L+\beta C+ \gamma R,
\end{equation}
where we empirically set $\alpha$ to $1$, $\beta$ to $15$, and $\gamma$ to $0.1$ which scales the three terms into similar weights.

\section{Experiments}
We report on a set of experiments conducted to assess the benefit of our model. Section~\ref{sec:baselines} describes the state-of-the-art (SOTA) methods, along with the different baselines we evaluated. Section~\ref{sec:details} gives the implementation details. Moreover, we compare our model with the baseline methods in Section~\ref{sec:evaluation}. Finally, Section~\ref{sec:errors} reports and analyses the errors of our model.


\subsection{Baselines}
\label{sec:baselines}

Our model was compared against the following baseline and SOTA methods. For training these baselines, we used all samples in training set, where for the ones with dynasty or century labels, the middle years were employed instead of Gaussian function-based representation.

\textbf{Nearest Neighbour Search}. We store feature vector $g_0(x)$ of $x$ and $t$ for $(x, t) \in \mathcal{D}_\text{L}$, where $g_0(x)$ is the feature embedding from a ResNet50 variant of ArcFace \cite{deng2019arcface} (the subscript 0 is to emphasise that it is the original pretrained model). For a new Buddha face image $x'$, we compute $g_0(x')$ and find the nearest neighbour from the stored feature vectors in terms of the cosine similarity. The prediction by this method is given as the label $t$ associated with the nearest neighbour.

\textbf{Gaussian Process Regression}. Gaussian process regression \cite{williams2006gaussian} is a non-parametric kernel-based probabilistic model. We used $g_0$ as our feature extractor and MATLAB implementation with default parameters. 

\textbf{ResNet50 Regression w/o fine-tuning}.
We evaluate the same network structure as ours but without fine-tuning the backbone. The model is trained solely by our MSE loss.

\textbf{Semi-supervised Deep Kernel Learning (SSDKL)} \cite{jeanxieermon_ssdkl_2018}. SSDKL was proposed by Jean et al.~\cite{jeanxieermon_ssdkl_2018} and utilises the power of deep learning and Gaussian process to learn a model from both labelled and unlabelled data. Their objective is to maximise the likelihood of labelled data and at the same time minimise the predictive variance in the unlabelled data. We used their original implementation.\footnote{\textcolor{blue}{\url{https://github.com/ermongroup/ssdkl}}}

\textbf{GCNBoost Regression}. A GCN-based transductive semi-supervised learning classifier in \cite{GCNBoost} is adapted to the task of built year prediction. This method requires pseudo labels for unlabelled samples, and we used our full model to compute those pseudo labels.  A knowledge graph is built by connecting samples that have the same built years. We adapted the GCN to make it an output layer a single scalar representing the built year. The MSE loss term is used to train the model in an end-to-end manner.

\subsection{Implementation Details}
\label{sec:details}

Our model and its variants were implemented on PyTorch. For training, a A100 GPU with 40G of RAM was used. The batch size was set to 256, the learning rate was set to 0.003, and Adam was used for optimisation. For the MSE loss, we standardised the built year (i.e., $\mu$'s in the training set to have zero mean and unit variance). From our preliminary experiments, we found that data augmentation by only random horizontal flipping leads to the best performance; therefore, we used this in all experiments.

 
 \subsection{Results}
\label{sec:evaluation}
We evaluated our method and its variants for ablation studies, as well as the baselines listed in Section \ref{sec:baselines}. The test set of the dataset contains 1,485 samples, where 547 of them have a built time label, and 371 have an exactly built year label. Our evaluation uses only these 371 samples as it is not trivial to evaluate errors based on dynasty and century labels. We employed mean absolute error (MAE) as our error metric. 

Table \ref{tab:algoperformance} shows the performances of all methods in comparison. Rows 1, 2, 3 and 6 are the inductive baselines. Nearest Neighbour Search (row 1) and Gaussian Process Regression (row 2) did not perform well, while ResNet50 Regression w/o fine-tuning (row 3) gave competitive performances. This implies that training with the MSE loss can benefit a lot, and our backbone provides rich cues about Buddha statues even though it is trained on human faces; yet end-to-end training helps. These results may also suggest that the relationship between the feature space and the built year space is not simple, which can support the use of our KL divergence-based loss and regularisation term. 

Rows 4 and 5 are the performances of transductive baselines.  GCNBoost Regression \cite{GCNBoost} shows a better performance when compared to SSDKL \cite{jeanxieermon_ssdkl_2018}, but both methods did not perform as well as other methods. For SSDKL, we consider that the original feature vector obtained from our backbone did not have sufficient cues about the built years and the error accumulated in the iterations. GCNBoost Regressor suffers from the sparse connectivity between samples in the test set and those in the training set since we added edges between the samples with exactly same built years; therefore, the features cannot be well trained. Moreover, we do not have Buddha statues attributes other than the built years ones. Those attributes (if available) can improve GCNBoost accuracy as shown in \cite{GCNBoost} as they emphasis the relationship between the nodes in GCNBoost's knowledge graph.   


Rows 6 to 10 show performances of our model on the ablation study over the loss terms, where ``MSE'', ``KL'', and ``Reg.'' stand for the MSE loss, KL divergence-based loss, and regularisation terms, respectively. The results clearly show the importance of the MSE loss term, while the KL divergence-based loss and regularisation terms provide less impact on the performance. This is quite reasonable because the MSE loss term is the only one that gives direct supervision about the built time, while the others only tell the relationships among the samples. Yet, the best performance is obtained when we combined all three terms, which implies their complementarity.

\begin{table}[t]
\caption{Comparison of different methods and some variants of ours. Only 371 samples in the test set that have exact built year labels were used.}
\label{tab:algoperformance}
\begin{tabular}{l l r r}
\toprule
     \textit{} & Methods & MAE (Year) \\

\midrule
1& Nearest Neighbour Search    & 130.9 $\pm$ 9.8               \\ 
2& Gaussian Process Regression & 199.9 $\pm$ 5.4                 \\ 
3& ResNet50 Regression w/o fine-tuning & 73.8 $\pm$ 4.0   \\ 

\midrule
4& GCNBoost Regression \cite{GCNBoost} & 217$\pm$ 15.5                 \\ 
5& SSDKL \cite{jeanxieermon_ssdkl_2018}                            & 245.3$\pm$ 4.0                  \\ 
\midrule
6& Ours (MSE) & 56.2 $\pm$ 3.7    \\ 
7& Ours (KL+Reg.)                           &  338.3 $\pm$  33.1   \\ 
8& Ours (MSE+KL)                           & \textit{40.2 $\pm$ 3.62}    \\
9& Ours (MSE+REG)                           & \textit{39.3 $\pm$ 3.55}    \\
10&  Ours (MSE+KL+Reg.)                           & \textbf{37.5$\pm$ 3.64 }                             \\ 
\bottomrule
\end{tabular}
\end{table}

For tasks that involve cultural heritage, the availability of samples (i.e., images in our case) is a critical problem. Cultural heritages are often stored in a secure place, and very limited access is allowed. Under this circumstance, the only way to acquire images of cultural heritages is to make digital scans of their images printed in catalogues or books, as Renoust et al.~did in \cite{renoust2019historical}. The image quality is thus affected by various factors, such as the quality of captured images, the quality of printing, and the quality of digital scans. Moreover, image quality degradation due to these factors may not necessarily be distinguishable from the texture of Buddha statues themselves. This is an inherent problem that particularly rises with cultural heritage. We therefore investigated the performance of our method with respect to the image quality.

\begin{figure}[t]
	\centering
	\subfloat{\includegraphics[width=6cm]{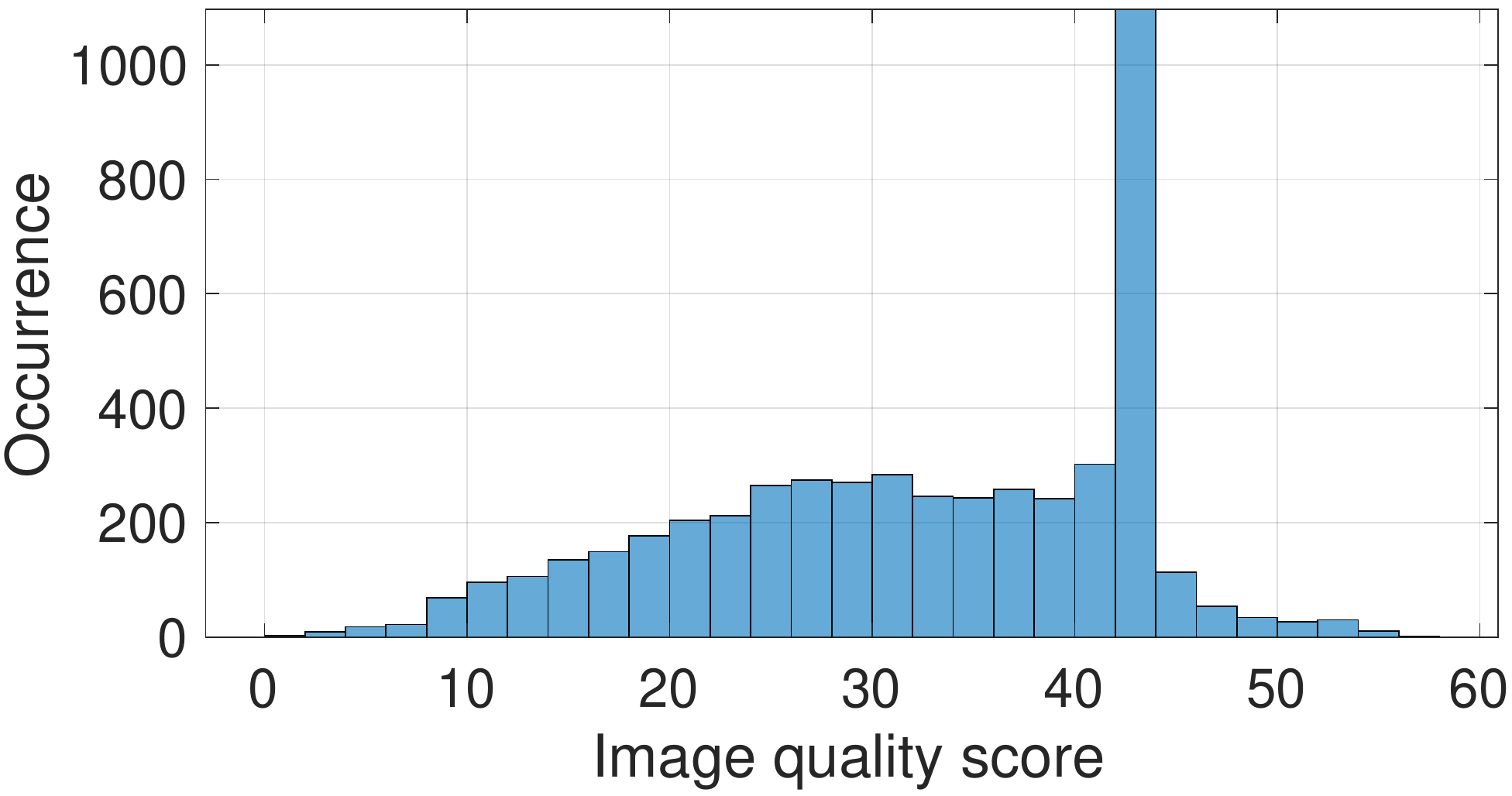}}\\
		\caption{The distribution of image quality scores measured by BRISQUE.}
	\label{fig:imageQDistribution}
\end{figure}

\begin{figure}[t]
	\centering
	\subfloat[0.56]{\includegraphics[width=2.6cm]{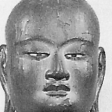}} \thinspace
	\subfloat[1.22]{\includegraphics[width=2.6cm]{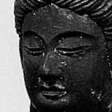}} \thinspace
	\subfloat[2.37]{\includegraphics[width=2.6cm]{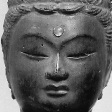}} \thinspace
	\\
	\subfloat[33.5]{\includegraphics[width=2.6cm]{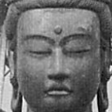}} \thinspace
	\subfloat[33.5]{\includegraphics[width=2.6cm]{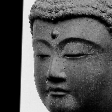}} \thinspace
	\subfloat[33.5]{\includegraphics[width=2.6cm]{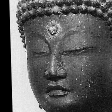}} \thinspace
	\\
	\subfloat[54.7]{\includegraphics[width=2.6cm]{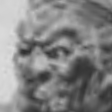}}\thinspace
	\subfloat[54.8]{\includegraphics[width=2.6cm]{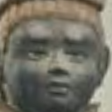}}\thinspace
	\subfloat[55.3]{\includegraphics[width=2.6cm]{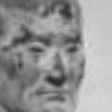}}\thinspace
	
	\caption{Example images with highest (a-c), middle (d-f) and lowest (g-i) quality scores.}
	\label{fig:imageQ}
\end{figure}

We employed Blind/Referenceless Image Spatial Quality Evaluator (BRISQUE) \cite{mittal2012no} to measure the image quality. BRISQUE uses an image quality score in $[0,100]$, where a lower score indicates better quality. The distribution of the image quality scores in the entire dataset (including both training and test sets) is shown in Fig.~\ref{fig:imageQDistribution}. The correlation between the image quality scores and built year labels is $-0.3021$, which weakly indicates that a statue captured with a higher image quality tends to be established more recently. Example images with the highest, medium, and lowest qualities are shown in Fig.~\ref{fig:imageQ}. The highest-quality images have a clear view of the Buddha face with less noises, while the medium ones show some noise or slight blur. The lowest-quality images suffer from a lower resolution (or severe blurring). 

\begin{figure}[t]
	\centering
	\subfloat{\includegraphics[width=6.5cm]{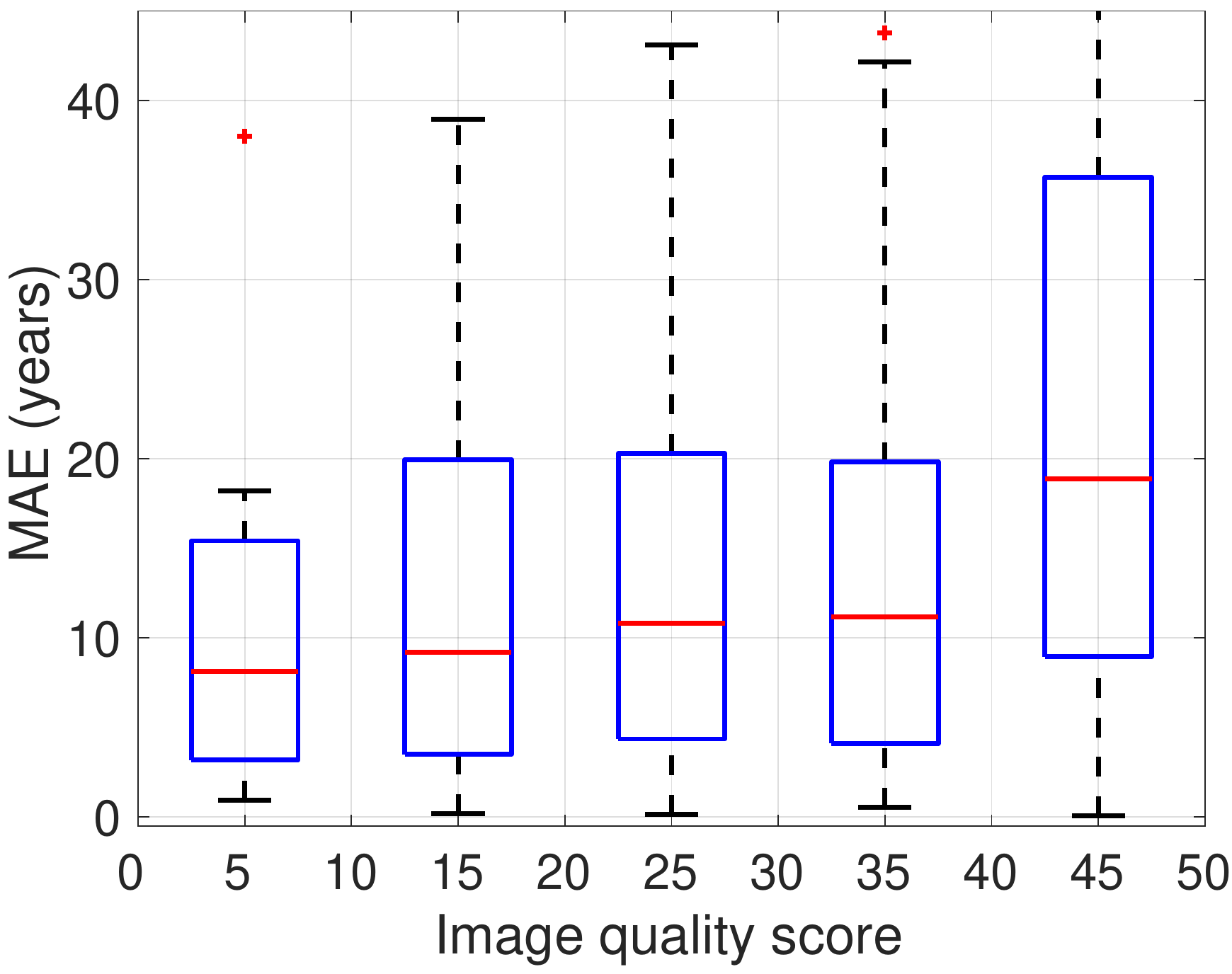}}
	\caption{The relationship between image quality scores and prediction errors in MAE.}
	\label{fig:image_quanlty_vs_error}
\end{figure}

To see the impact of the image quality on our method, we plotted the relationship between the image quality score and prediction errors in Fig.~\ref{fig:image_quanlty_vs_error}. The figure shows that images with a higher quality (i.e., a lower score) got a lower error, and vice versa. From this,  we can conclude that the image quality is an important factor to determine the prediction accuracy.

\subsection{Discussion}

\subsubsection{With conditional probability $q(f(x')|f(x))$ and $p(t'|t)$}

\begin{figure}[t]
	\centering
	\subfloat[$p(t'|t)$, $q(f(x')|f(x))$]{\includegraphics[width=4.2cm]{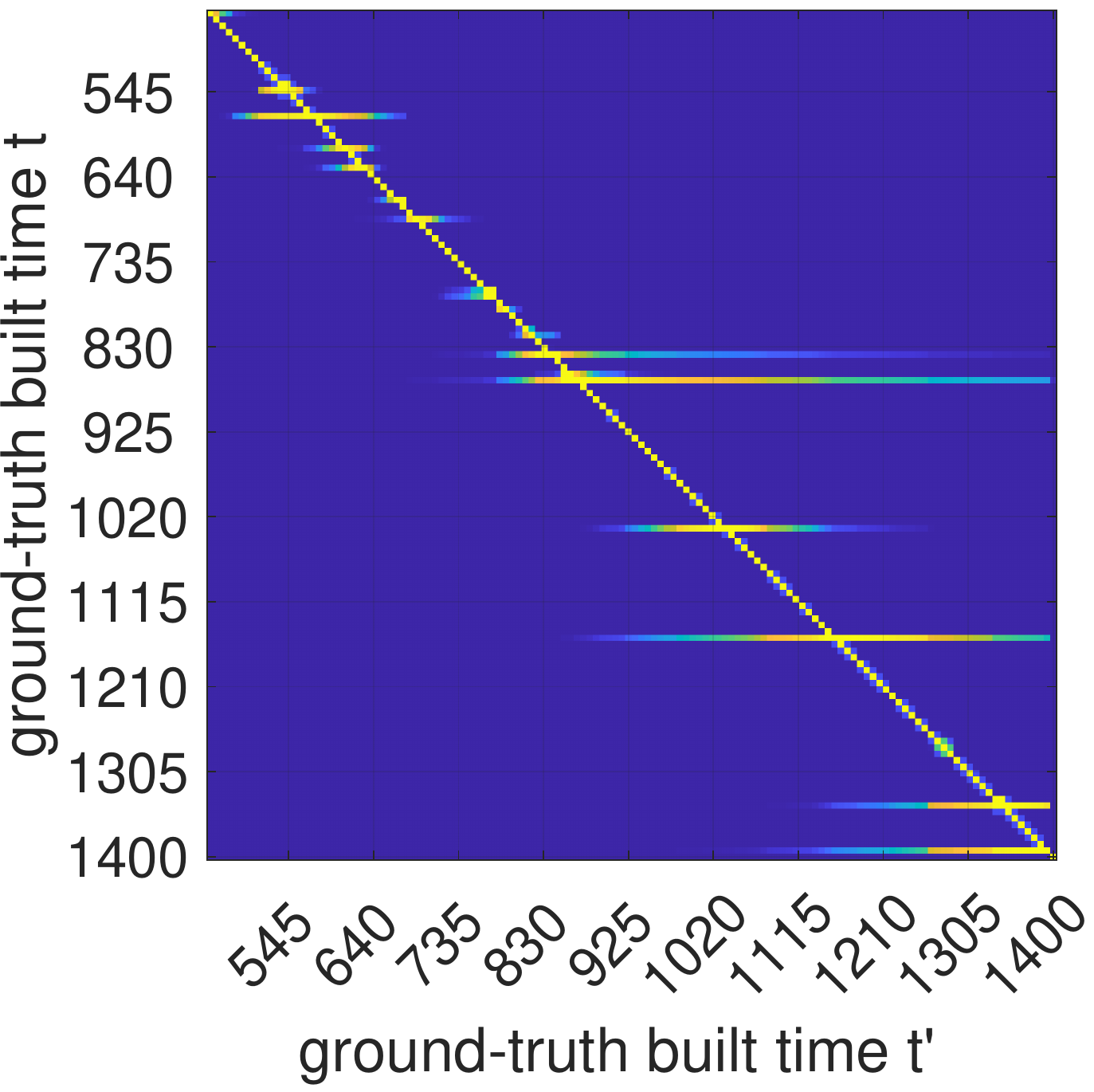}}
	\subfloat[MSE+KL]{\includegraphics[width=4.2cm]{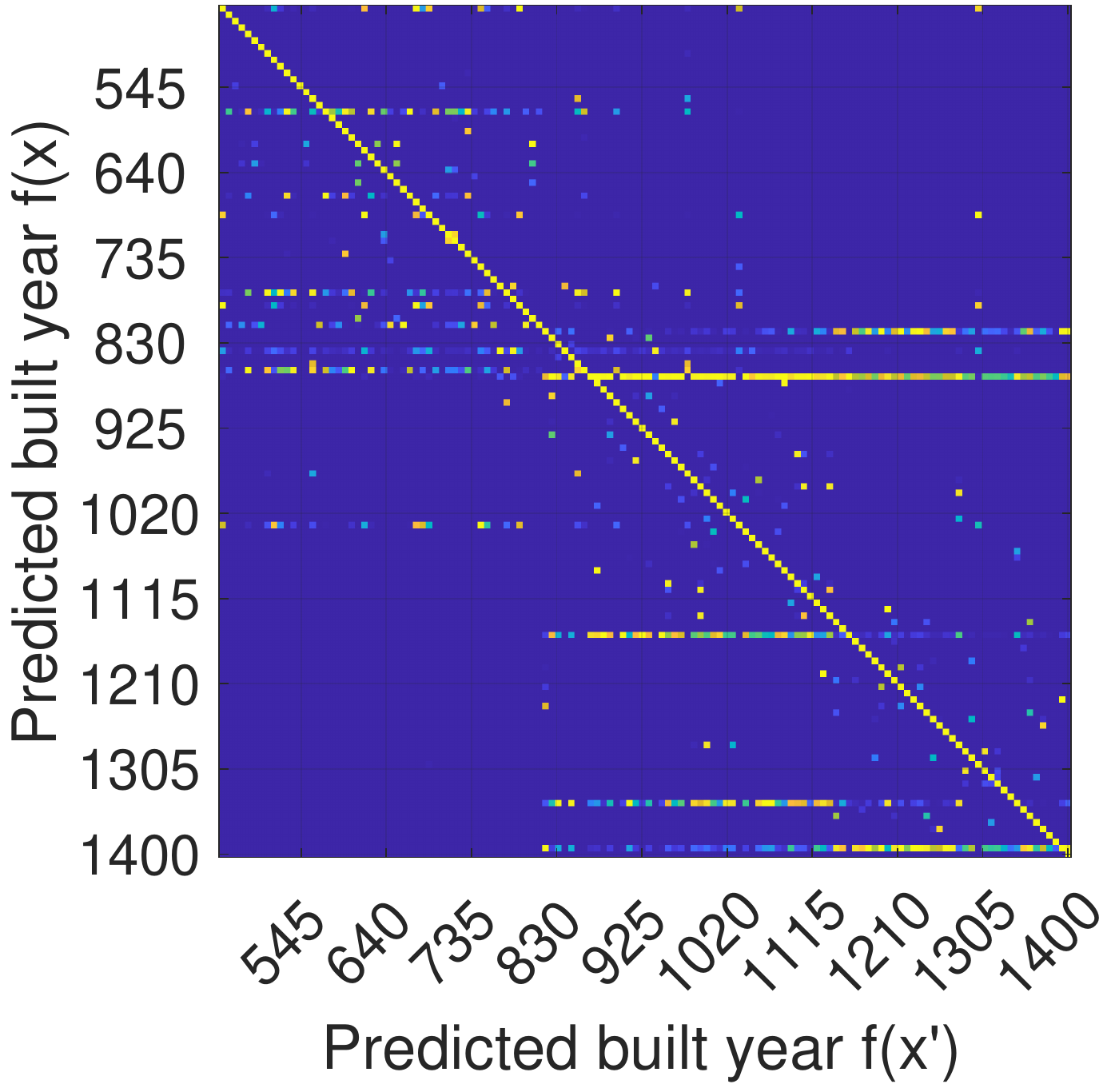}}\\
	\subfloat[MSE+Reg.]{\includegraphics[width=4.2cm]{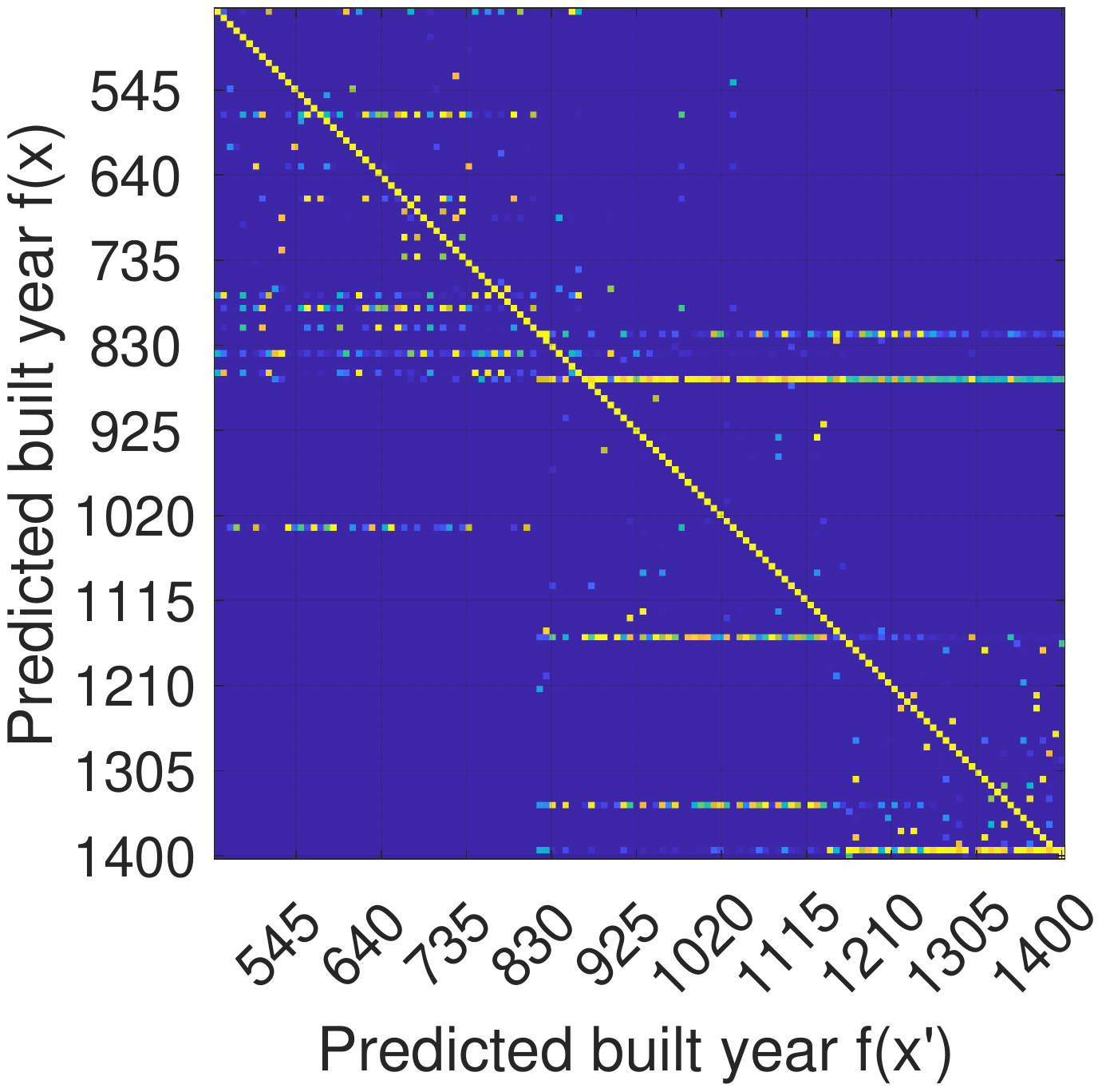}}
	\subfloat[MSE+KL+Reg.]{\includegraphics[width=4.2cm]{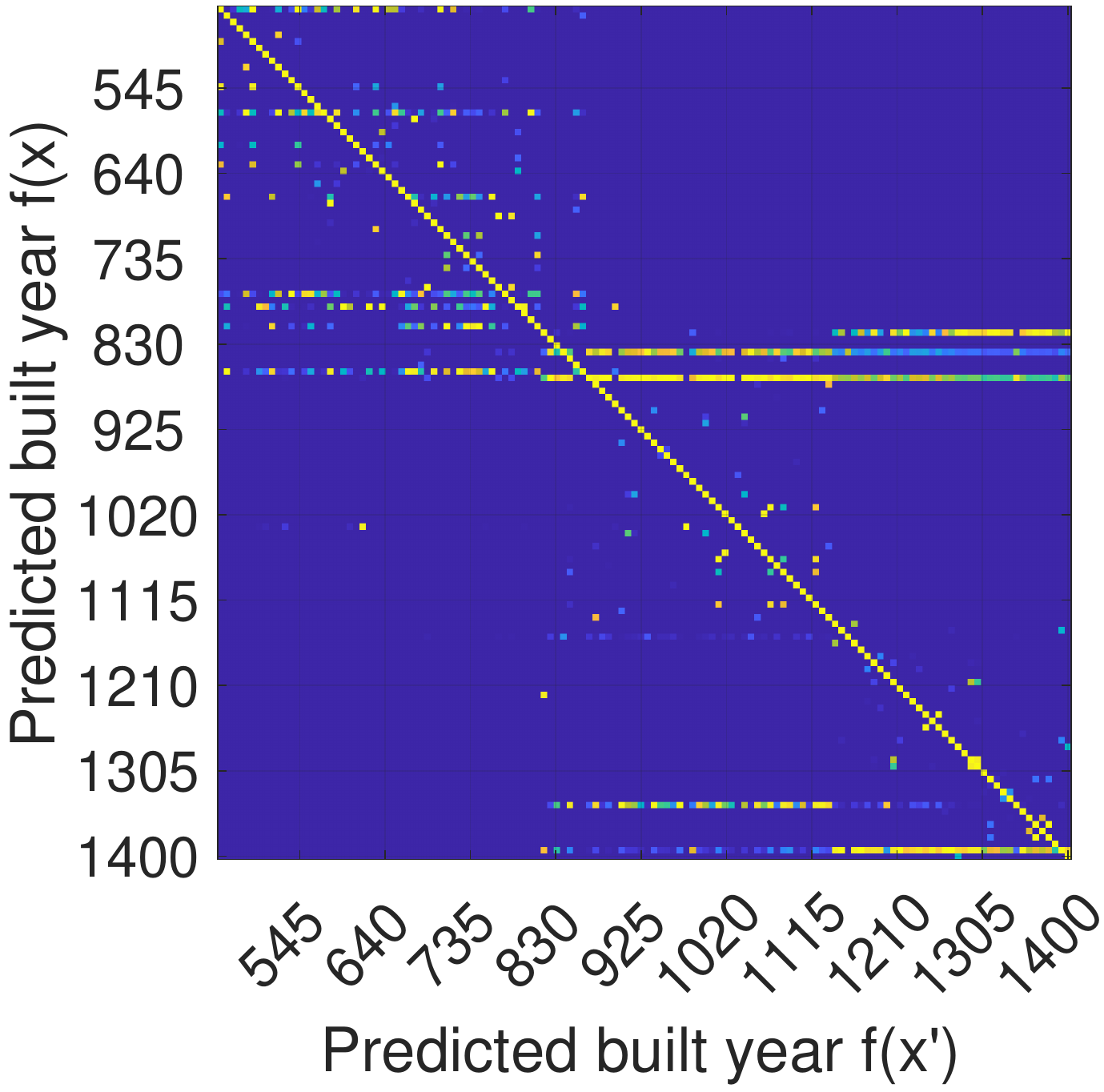}}
	\caption{Visualisation of conditional probabilities (a) $p(t'|t)$, $q(f(x')|f(x))$ for (b) MSE+KL, (c) MSE+Reg., and (d) MSE+KL+Reg.}
	\label{fig:conditionalprobabilities}
\end{figure}

The conditional probabilities $q(f(x')|f(x))$ and $p(t'|t)$, which are used for our KL divergence-based loss term, encode the proximity between a pair of built times (either of the prediction and ground truth labels). This can be informative to comprehend the effect of our loss terms. We visualise $q(f(x')|f(x))$ and $p(t'|t)$ in Fig.~\ref{fig:conditionalprobabilities}. For $q(f(x')|f(x))$, we generated three visualisations for different combinations of loss terms, i.e., MSE+KL, MSE+Reg., and MSE+KL+Reg. (our full model).  The yellow highlight of the built years indicates that the corresponding predictions ($f(x)$ and $f(x')$) or ground-truth labels ($t$ and $t'$) are close to each other. The conditional probabilities are computed over the test set, where the samples are in the chronological order based on the ground-truth labels (for dynasty and century labels, we take their middle years). Therefore, in the ideal case where an exact built year label is assigned to all samples, the highlight forms the diagonal line; however, as there are dynasty and century labels, we observe horizontal lines.  


Comparisons between Fig.~\ref{fig:conditionalprobabilities} (b) versus (c) and (d) demonstrate the importance of the regularisation term $R$. The regularisation term has two benefits: First it enriches our training process with large amount of unlabelled data. Second, it acts as a smoothing function that push the samples with similar properties closer in the manifold. Figure \ref{fig:conditionalprobabilities} (b) shows a more scattered distribution. On the other hand, Fig.~\ref{fig:conditionalprobabilities} (c) shows a greater amount of estimation concentrated in a narrower band where its distribution shares more similarity with ground truth map.

\begin{figure}[t]
	\centering
	\subfloat{\includegraphics[width=0.5\textwidth]{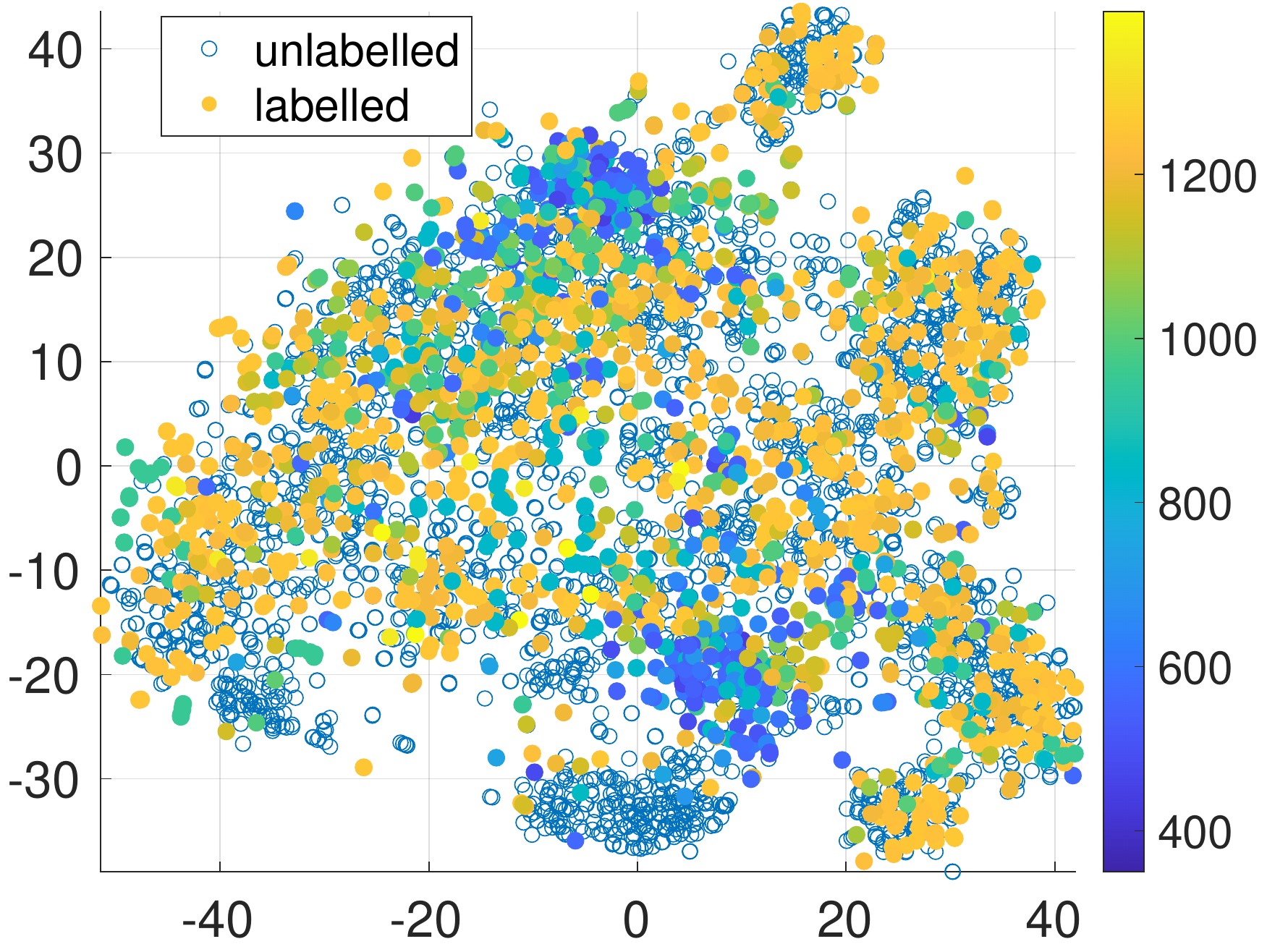}}
	\caption{The t-SNE visualisation on the feature vectors before fine-tuning.}
	\label{fig:visualizationbefore}
\end{figure}

\begin{figure}[t]
	\centering
	\subfloat{\includegraphics[width=0.5\textwidth]{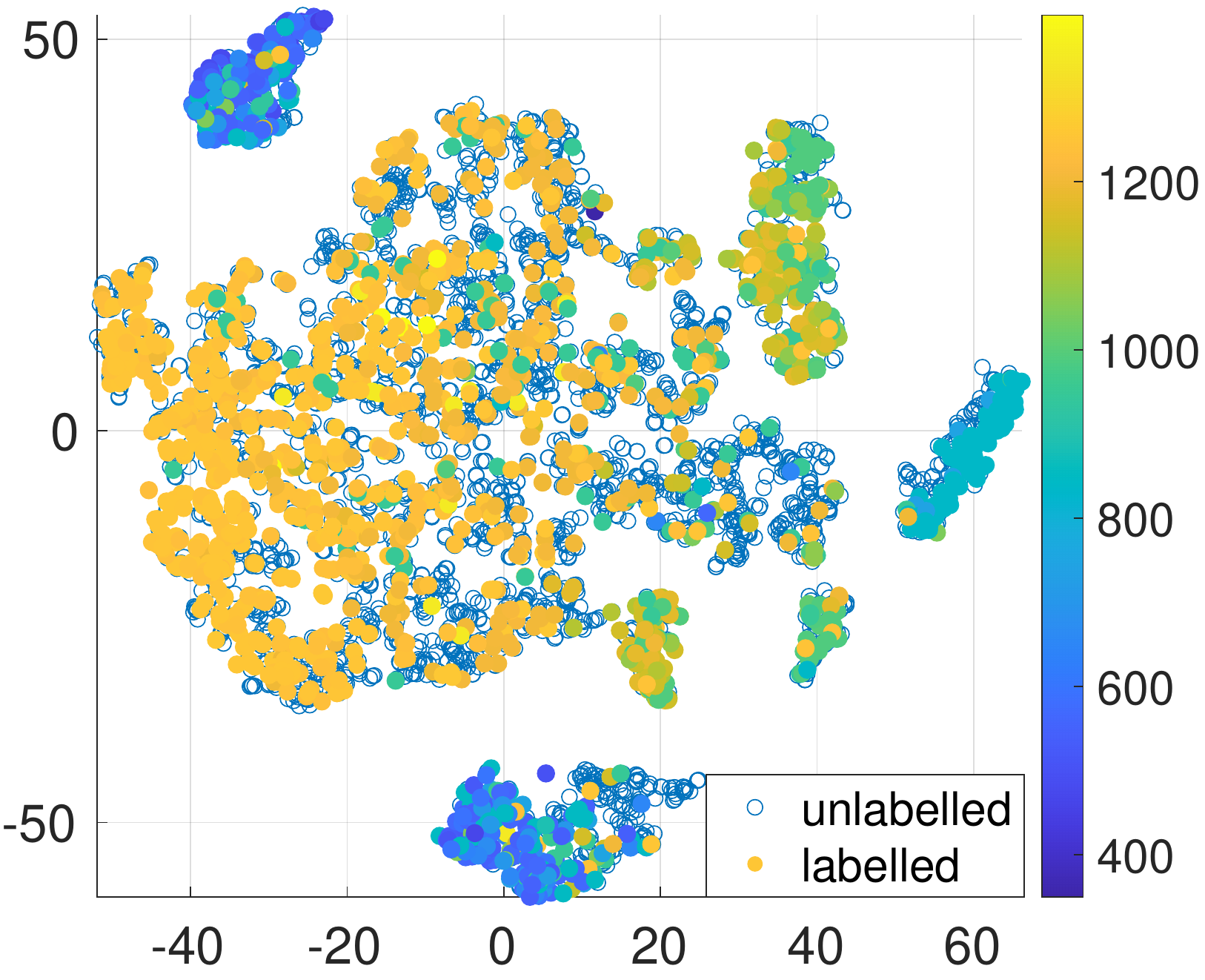}}
	\caption{The t-SNE visualisation on the feature vectors after fine-tuning.}
	\label{fig:visualizationafter}
\end{figure}

The t-SNE visualisations of the feature vector extracted from both training and test sets before and after fine-tuning are shown in Figs. \ref{fig:visualizationbefore} and \ref{fig:visualizationafter}, respectively. For the labelled samples, the built years (the middle years for dynasty and century labels) are colour coded. Before fine-tuning, we can see that the year is almost randomly distributed. Whereas after fine-tuning, the distribution of the samples looks to be more structured, which indicates that our loss function effectively gives supervision to the backbone based on the similarity in the feature vectors and ground-truth labels.

\subsubsection{Qualitative analysis}
\label{sec:errors}

\begin{figure}[htpb!]
	\centering
	\subfloat[error=0]{\includegraphics[width=2.6cm]{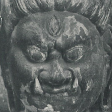}}\thinspace 
	\subfloat[error=0]{\includegraphics[width=2.6cm]{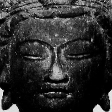}}\thinspace 
	\subfloat[error=0]{\includegraphics[width=2.6cm]{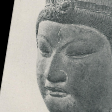}}\\ 
	\subfloat[error=16]{\includegraphics[width=2.6cm]{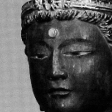}}\thinspace
	\subfloat[error=16]{\includegraphics[width=2.6cm]{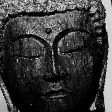}}\thinspace 
	\subfloat[error=16]{\includegraphics[width=2.6cm]{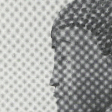}}\\ 
	\subfloat[error=444]{\includegraphics[width=2.6cm]{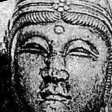}}\thinspace 
	\subfloat[error=469]{\includegraphics[width=2.6cm]{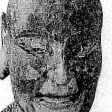}}\thinspace 
	\subfloat[error=601]{\includegraphics[width=2.6cm]{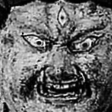}}\\ 
	\caption{Examples of images with low (first row), medium (second row) and high error (third row) estimation error in our test set.}
	\label{fig:failuremode}
\end{figure}

The samples with low, medium and high prediction errors in the test set are shown in Fig.~\ref{fig:failuremode}. The statues with low and medium errors contain clear facial features, while ones with high errors tend to have more visible damage and lower image quality. The statues with high errors were built in 1091, 1047, and 1241, respectively; however, our method predicted them as 647, 578, and 640. The source of the errors can come from the low image quality; for example, image (g) looks to have block noises due to compression, while (h) and (i) are dark and blurry. This is consistent with our result in Section \ref{sec:evaluation}.

\section{Conclusion}
In this work, we proposed a method to estimate the built year of Buddha statues from only their face images. We faced four main challenges to solve this task: First, the images in the dataset were collected from digital scans of 5 different books, which introduced many artefacts and distortions. Second, the dataset is quite small, containing only 4,949 facial images. Third, only 30\% of samples in this small dataset are labelled. Finally, this dataset contains two types of labels: exact built years and the range of built years (i.e, dynasty and centuries) estimated by historians. To overcome those challenges, we modelled the labels in the form of Gaussian functions, which provided a unified built time representation for training. We also designed a new loss function to handle Gaussian function-based built time representation as well as unlabelled samples. Our experimental results showed that our method outperformed state-of-art methods and baselines by a significant margin. As our future work, we plan to use the different Buddha statues related information available in the dataset \cite{renoust2019historical}, such as built material, original location, which can be correlated with built time. For this, we need to handle highly heterogeneous data as the dataset has a lot of missing entries. The second direction is to collect more samples for better training.


\bibliographystyle{ACM-Reference-Format}
\bibliography{sample-base}

\end{document}